\documentclass[a4paper,twoside]{article}

\usepackage{gensymb}
\usepackage{epsfig}
\usepackage{subfigure}
\usepackage{calc}
\usepackage{amssymb}
\usepackage{amstext}
\usepackage{amsmath}
\usepackage{amsthm}
\usepackage{multicol}
\usepackage{pslatex}
\usepackage{apalike}
\usepackage{tabularx}
\usepackage{graphicx}
\usepackage{adjustbox}
\usepackage{SCITEPRESS}     

\input{alphabet.tex}
\input{abrege.tex}

\begin{document}

\title{Discriminant Patch Representation for RGB-D Face Recognition Using Convolutional Neural Networks}

 \author{\authorname{Nesrine Grati\sup{1}, Achraf Ben-Hamadou\sup{2} and Mohamed Hammami\sup{1}}
 \affiliation{\sup{1}MIRACL-FS, Sfax University, Road Sokra Km 3 BP 802, 3018 Sfax, Tunisia}
 \affiliation{\sup{2}Digital Research Center of Sfax, Technopark of Sfax, PO Box 275, 3021 Sfax, Tunisia}
 \email{\ grati.nesrine@gmail.com, achraf.benhamadou@crns.rnrt.tn\, mohamed.hammami@fss.rnu.tn}
 }

\abstract{This paper focuses on designing data-driven models to learn a discriminant representation space for face recognition using RGB-D data. Unlike hand-crafted representations, learned models can extract and organize the discriminant information from the data, and can automatically adapt to build new compute vision applications faster. We proposed an effective way to train Convolutional Neural Networks to learn face patch discriminant features. The proposed solution was tested and validated on state-of-the-art RGB-D datasets and showed competitive and promising results relatively to standard hand-crafted feature extractors.}

\onecolumn \maketitle \normalsize \vfill
\section{\uppercase{Introduction}}
\label{sec:introduction}

With the increase demand of robust security systems in real-life applications, several automated biometrics systems for person identity recognition are developed, where the most user-friendly  and non-invasive modality is the face. Face recognition using 2D images was well treated but still affected by imaging conditions. Thanks to the 3D technology progress, the recent research has shifted from 2D to 3D \cite{bowyer2006survey,abbad20183d}. Indeed, three-dimensional face representation ensures a reliable surface shape description and add geometric shape information to the face appearance. Most recently, some researchers used image and depth data capture from low-cost RGB-D sensors like MS Kinect or Asus Xtion instead of bulky and expensive 3D scanners. In addition to color images, RGB-D sensors provide depth maps describing the scene 3D shape by active vision. With the availability of cost-effective RGB-D sensors, many researchers proposed and adapted feature extraction operators to the raw data for different computer vision applications like gait analysis \cite{wu2012one}, lips movement analysis \cite{Rekik2016,rekik_unified_2015,rekik_human_2015}, and gender recognition \cite{huynh2012efficient}. Hand-crafted or engineered feature extractors such as LBP, Local Phase Quantization(LPQ), HOG were mainly used to deal with RGB-D data for face recognition. The main benefits of these feature extractors is that they are relatively simple and efficient to compute. Alternatively, learned features, for example with Convolution Neural Networks (CNNs), achieve a very prominent performance in many computer vision tasks (\eg object detection \cite{szegedy2013deep}, image classification \cite{krizhevsky2012imagenet} \etc). The basic idea behind is to learn data-driven models that transform the raw data to an optimal representation space leading to appropriate features without manual intervention.

In this context, this paper focuses on the feature extraction part in our face recognition pipeline. A given face is represented by a set of patches extracted from image and depth data. We propose to learn discriminant local features using data-driven representation to describe the face patches before feeding a Sparse Representation Classification (SRC) algorithm to attribute the person identity. 

The rest of this paper is organized as follows. First, an overview on the most prominent {RGB-D} face recognition systems is given in section 2. Then, we detail our proposed system in section 3. Section 4 summarizes the performed experiments and the obtained results to validate our proposed system. Finally, we conclude this study in section 5 with some observations and perspectives for future work.

\section{\uppercase{Related Works}}

In this overview, we focus mainly on the feature extractors for face recognition using RGB-D sensors. Actually, many other aspects can be discussed like the pre-processing techniques, or the overall classification schemes. 
In \cite{li2013using}, the nose tip is manually detected and the facial scans are aligned with a generic face model using the Iterative Closest Point (ICP) algorithm to normalize the head orientation and generate a canonical frontal view for both image and depth data. A symmetric filling process is applied on the missing depth data specifically for the non frontal view. For image data, Discriminant Color Space (DCS) operator is used as feature extractor. Then, obtained depth frontal view and DCS features are classified separately using SRC before late fusion to get the person identity.

\cite{hsu2014rgb} fits a 3D face model to the face data to reconstruct a single 3D textured face model for each person in the gallery. The approach requires to estimate the pose for any new probe to be able to apply it to all 3D textured models in the gallery. This allows to generate 2D images by plan projection and then compute the LBP descriptor on the whole projected 2D images to perform the classification using an SRC algorithm. Likewise, \cite{sang2016pose} used the depth data for pose correction based on ICP algorithm to render the gallery view as the probe one. However, contrary to \cite{hsu2014rgb}, the authors applied Joint Bayesian Classifier on RGB and depth HOG descriptors extracted from the both data and the final decision is made via weighted sum of their similarity scores.

From the discussion above, the most focus to pre-processing especially dealing with pose variation by aligning the query data to the gallery samples. Although this kind of sequential processing may lead to error propagation from pose estimation to the classification, it gave a very good results \cite{hsu2014rgb}. Alternatively, to deal with pose variation, \cite{ciaccio2013face} used a large number of image sets in the gallery under different poses angles from a single RGB-D data. Also, the face pose is estimated via the detection and alignment of standard facial landmarks in the images \cite{zhu2012face}. Each face is then represented using a set of extracted patches centered on the detected landmarks and described by a set of LBP descriptor, co-variance of edge orientation, and pixel location and intensity derivative. The classification is then performed by computing distances between patch descriptors, inferring probabilities, and lately performing a Bayesian decision. 

The following works, gave more attention to feature extraction from face RGB-D data than pre-processing and dealing with head pose variation. In \cite{dai2015multi}, a single ELMDP (Enhanced Local Mixed Derivative Pattern) descriptors are extracted and Nearest Neighbor algorithm is used for the combined features with confidence weights. In \cite{goswami2014rgb}, a combination of HOG applied on saliency and entropy features, and geometric attributes computed from the Euclidean distances between face landmarks are used as face signature. The random forest classifier is then used for the identity classification. In \cite{boutellaa2015use}, a series of hand-crafted feature extractors (\ie LBP, LPQ, and HOG) are applied respectively on texture and depth crops and finally SVM classifier is carried out for face identification. The only use of the carefully-engineered representation was with feature Binarized Statistical Image Features detector (BSIF).

In \cite{kaashki2018rgb} three-dimensional constrained local model (CLM-Z) is applied for face-modeling and landmarks points localization. Local features HOG, LBP and 3DLBP around landmarks points are extracted then SVM classifier is used for the classification.

Indeed, \cite{hayat2016rgb} proposed the first RGB-D image set classification for face recognition. For a given set of images (which can captured frames with Kinect sensor), the face regions and the head poses are firstly detected with \cite{fanelli2011real} algorithm's than clustered into multiple subsets according to the estimated pose. A block based covariance matrix representation with LBP features is applied to model each subsets on the Riemannian manifold space and SVM classification is performed on all subsets for the both modality. The final decision is made with a majority vote fusion. The proposed technique has been evaluated on a combination of three RGB-D data sets and achieved an identification rate of 94.73\% which concurrent the single image based classification accuracy's.

\begin{figure*}[!h]
   \centering
   {\epsfig{file = 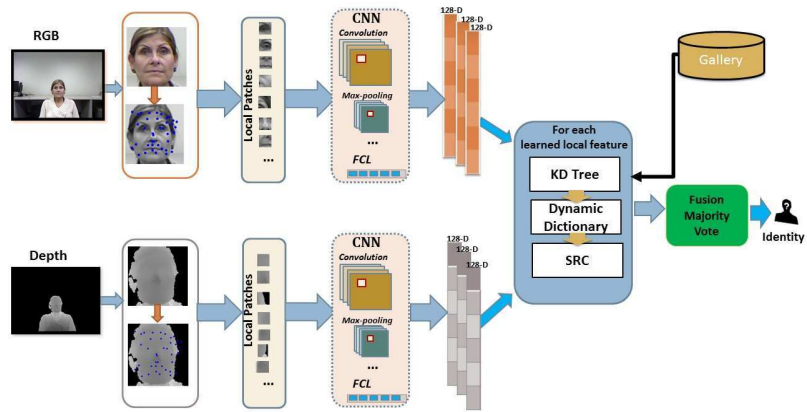, width = 10.5cm}}
  \caption{ An overview of the online process of the proposed method. First, the facial regions are detected from image and depth data. Local patches are then extracted from both of the modalities. Afterward, two CNNs are used to transform the extracted patches to obtain feature vectors to feed the SRC algorithm and finally obtain the person ID.
}
  \label{fig:flowchart}
 \end{figure*}

\paragraph{Observations:}

For the classification part, we observe that SRC is used in the most popular RGB-D face recognition systems \cite{li2013using,hsu2014rgb}. Indeed, after its successful application in \cite{wright2009robust} for face recognition, SRC has attracted the attention in many other computer vision tasks. Also, SRC is a good choice when the number of classes in the dataset is variable and in a constant increase, which is the case of face recognition applications.

We note that the most prominent methods \cite{hsu2014rgb} \cite{li2013using} aim principally to overcome the issues of pose variation either with pose correction or with augmenting the gallery through generating new images in different view or even capturing multi-view data for each face in the gallery. Data representation and appropriate feature extraction remains underestimated in the aforementioned works and they settle for hand-crafted features combinations and fusion. In this new era of deep learning techniques, we believe that {RGB-D} face recognition systems can take benefits of CNNs to learn appropriate features and boost their performances.

\paragraph{Contribution:} 

The main objective of this paper is to highlight how CNNs can contribute to learn a discriminant representation of local regions in face image, and how it competes with standard hand-crafted feature extractors in the case of RGB-D face recognition. A given face is represented by a set of patches around detected salient key-points. Each of these patches is assigned an ID by SRC technique that associate the patch to one of the most similar patches in the database. The raw patches data (image and depth) are transformed using a CNN before feeding the classification part. We propose an effective approach to learn our CNN weights leading to a discriminant space for face patches representation.

 \section{\uppercase{Proposed face recognition system}}

The proposed approach involves online and offline phases sharing some processing blocks. The offline phase is to train our CNNs while the online phase is dedicated to predict the person identity given a face query. Figure \ref{fig:flowchart} sketches the online phase. It goes along the following steps. Firstly, the face is localized in the image. It is then represented by a set of patches cropped around key-points extracted on the face. Afterward, two trained CNNs are applied to transform these patches to get a feature vector for each one, and an SRC algorithm is used to attribute an ID for each feature vector before making the late fusion leading to the predicted identity. The remaining of the section gives details about the different processing blocks just introduced including the training of the CNNs.

\subsection{Face pre-processing and patch extraction}

The face pre-processing shared between the offline and online phase of our system includes median and bilateral filtering for the depth maps and face localization \cite{zhu2012face}\footnote{We used only the face localization, facial landmarks were not used.}. The face region is cropped and resized to $96\times96$ pixels to ensure a normalized face spatial resolution. To get rid of face landmarks localization, we only consider salient image key-points without any further semantic analysis and without loss of generality. In other words, we do not try to catch specific anatomical reference points on the face. That said, the repeatability of image feature points for face analysis was proven. We used the SURF operator \cite{bay2006surf} to extract interest key-points on the cropped face images. The number of extracted key-points is variable and depends on face textures and also the position of the person in the frustum of the {RGB-D} camera. The key-points coordinates are mapped on the depth crop using the sensor calibration parameters. Around each key-point, we crop from both image and depth data two patches of $20\times 20$ pixels. Again, the mapping between image and depth map can be ensured by the RGB-D sensor geometric calibration \cite{ben-hamadou_flexible_2013}.
\subsection{CNN architecture and training}

Since the size of the CNN input patches is small ({20$\times$20 pixels}), we designed a relatively shallow CNN architecture as described in Table \ref{tab:cnn_architecture}. It is worth to notice that at this level of the study we did not try complex architectures or fancy connectivity (skip connections, residual, \etc.) but it could be explored later in future works. We train separate models with the same architecture for each modality (\ie image and depth patches). 


\begin{table}[b]
\caption{Our CNN architecture for small RGB-D Patches. \textit{odim} stands for number of channels in the output tensors, similarly \textit{idim} is the number of channels in the input tensors, and \textit{ks} is the kernel size.} \label{tab:cnn_architecture} \centering
\small{
\begin{tabular}{|c|c|c|}
  \hline
  Layer & Parameters & Output tensor  \\
  \hline
  Convolution & odim: 6, \textit{ks}: 3 & (6,18,18)\\
  \hline
  BatchNorm & & (6,18,18) \\
  \hline
  Sigmoid &  & (6,18,18) \\
  \hline
  Max Pooling & \textit{ks}: 2 & (6,9,9)\\
  \hline
   Convolution & \textit{odim}: 32, \textit{ks}: 3 & (32,7,7)\\
  \hline
   ReLU & & (32,7,7) \\
  \hline
  Max Pooling & \textit{ks}: 2 & (32,3,3)\\
  \hline
  \small{FCL} & \textit{idim}: 288, \textit{odim}: 128 & 128\\
  \hline

\end{tabular}
}
\end{table}

Research on face analysis usually focuses on finding an improvement in faithful face characterization with discriminant and robust representation. Learning descriptors with neural networks is entirely a data-driven approach. The objective of the discriminant descriptors learning is to find a transformation from raw space to an another space in which features from same classes are closer than features from different classes. Metric learning using a triplet network was introduced by Google's FaceNet \cite{schroff2015facenet}, where a triplet-loss is used to train an embedding space for face image using online triplet mining which outperforms a Siamese networks in manifolds clustering. Good face embedding satisfy similarity's constraint by the way faces from the same person should be close together while those of different faces are far away from each other. The intra-class distance should to be smaller than the inter-class distance and form well separated clusters.

\begin{figure}
  \centering
   {\epsfig{file = 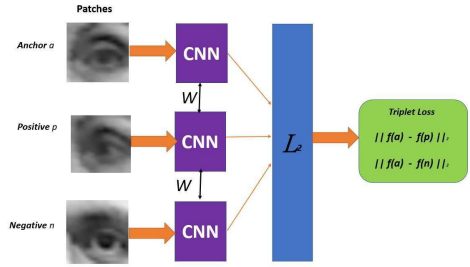, width = 7cm}}
  \caption{Local feature descriptor training pipeline with triplet loss.}
  \label{fig:tripletloss}

 \end{figure}
 
In this paper, the triplet loss takes a triplet of patches as input in the form $\ {a,p,n} $ , where$\ a $ is the anchor patch, $\ p $ is the positive patch, which is a different sample of the same class as $\ a $, and $\ n $ standing for negative patch is a sample belonging to a different class. The objective of the optimization process, is to update the parameters of the network in such way that patches $\ a $ and $\ p $ become closer in the embedded feature space, and $\ a $ and $\ n $ are further apart in terms of their Euclidean distances as presented in Figure \ref{fig:tripletloss}. The triplet loss formula is given in Equation \ref{eq1}. $\ f(x) $ stands for the application of CNN on a given input $x$ to generate a feature vector (embedding). Another hyper parameter is added to the loss equation called the margin, it defines how far away the dissimilarities should be.
Minimizing $\ L_{tr}$ enforces to maximize the Euclidean distance between patches from different classes which should be greater than the distance between anchor and positive features distance. For efficient training, only the triplets patches that verify the constraint $ \ L_{tr} > 0 $ are online selected as a valid triplet to improve the training.

\begin{eqnarray}
\label{eq1}
L_{tr} &=&  \sum_{\substack{a,n,p }}\left(||f(a)-f(p)||_2^2 \right.\\
& & -||f(a)-f(n)||_2^2  \nonumber\\
& & + \left. margin \right) \nonumber
\end{eqnarray}

\subsection{Patch classification}

We followed \cite{grati2016scalable} for the adaptive and dynamic dictionary selection in the SRC process. The objective the SRC is to reconstruct an input signal by a linear combination of atoms in a selected dictionary. 
We denote by $\yb_k \in \mathbb{R}^{M}$ the input feature vector obtained from the application of the trained CNN on the $k$-th extracted patch and $M$ is the its dimension. Also, we note by $\tilde{\Db_k} \in \mathbb{R}^{M \times \tilde{N}}$ the dictionary. It consists of the closest $ \tilde{N} $ gallery patches (atoms). Equation \ref{eq3} gives the linear regression leading to the reconstructed feature vector $\tilde{\yb_k}$.
$\xb_k \in \mathbb{R}^{\tilde{N}}$ is the sparse coefficient vector whose nonzero values are related to the atoms in $\tilde{D_k}$ used for the reconstruction of $\yb_k$, $\epsilon_k$ captures noise, and $\tilde{N}$ is experimentally fixed.

\begin{equation} \label{eq3}
    \tilde{\yb_k}= \tilde{\Db_k} ~ \xb_k+ \epsilon_k 
\end{equation}

The estimation of the sparse coefficients $\xb_k$ is formulated by a LASSO problem with an $\ell_1$ minimization using \cite{mairal2010online}:

\begin{equation}\label{eq4}
\arg min (   \| \tilde \Db_k~\  x_k-\yb_k \|_{2}+ \lambda \| \ \xb_k \|_{1})
\end{equation}
Finally, the identity associated to the $k$-th patch is classically the class generating the less reconstruction error. Once the sparse representation of all local patches in the query image is obtained, a score level fusion strategy is applied then a majority vote rule predicts the final person identity.


\section{\uppercase{Experiments and Results}}
\subsection{Training details}
Our CNNs has been trained with a batch size of 64, a decay value of $0.0005$ a momentum value of $0.09$ and an initial learning rate set to $0.001$. We used PyTorch \footnote{https://pytorch.org/} framework to implement and train the CNNs. Firstly, the set of patches is classically split to training and testing sets. The pool of patch triplets needed for the CNNs training are generated and updated every ten epochs of the training by gathering the triplets from all the persons equally. A single patch triplet is obtained following these 3 steps:
\begin{enumerate}
    \item Randomly select one anchor patch from the pool of patches related the a given person $c$.
    \item Randomly select the positive patch from the remaining patches in the same pool. 
    \item Randomly select the negative patch from the patch pool related to other persons ($\neq c$)
\end{enumerate}

\subsection{Datasets}
Our approach is validated on two publicly {RGB-D} face databases: Eurecom \cite{huynh2012efficient}, and Curtin faces \cite{li2013using}.\\
\begin{itemize}
    \item Eurecom dataset is composed by 52 subjects, 14 females and 38 males. Each person has a set of 9 images in two different sessions. Each session contains 9 settings: neutral, smiling, open mouth, illumination variation, left end right profile, occlusion on the eyes, occlusion on the mouth, and finally occlusion with a white paper-sheet. 
    \item CurtinFaces dataset consists of 52 subjects. Each subject has 97 images captured under different variations: combinations of 7 facial expression, 7 poses, 5 illuminations, and 2 occlusions. CurtinFaces database with low quality face models is more challenging in terms of variations of poses, and expression and illumination face models.
   
\end{itemize}

\subsection{Validation and results}

\begin{figure}[t]
  \centering
   {\epsfig{file = 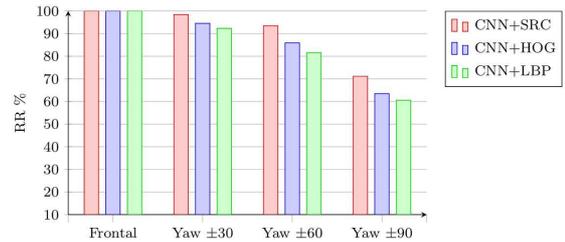, width = 7.5cm}}
  \caption{Performance comparison between our CNN features and the standard hand-crafted features HOG and LBP.}
  \label{fig:handcraftedvsLearned}
 \end{figure}

We performed two sets of experiments to evaluate our approach. The first set is dedicated to compare with state-of-the-art systems, and the second set is to evaluate the proposed learned features comparing to hand-crafted features (\ie HOG and LBP) taking our face recognition system as baseline. 

For the first set of experiments, and on CurtinFaces dataset, we report our results as well as those of \cite{hsu2014rgb} \cite{li2013using}, \cite{ciaccio2013face}, \cite{kaashki2018rgb}, and \cite{grati2016scalable}.
To make a fair comparison, we use the same protocol as \cite{li2013using}. 18 images containing only one kind of variations in illumination, pose or expression are selected for the training and testing images include the rest of non-occluded faces, there are a total of 6 different yaw poses with 6 different expressions added to the neutral frontal views. The reported results of our system on RGB, depth and fusion scheme in comparison with \cite{li2013using} are summarized in table \ref{tab:comp_our_li}. 
\begin{table}[h]

\centering
\caption{Face recognition performance under yaw pose and facial expression variations.}
\label{tab:comp_our_li}
\small{
\begin{tabular}{|c|c|c||c|}

\hline
Pose& Modality  & Our Work & DSC+SRC
\\\hline
Frontal & RGB  & 100\%& 100\%
\\\cline{2-4} & Depth &100\% &100\%
\\\cline{2-4}&Fusion& 100\% &100\%\\ 
\hline
\hline
Yaw $\pm 30$\degree  & RGB&99.03 \%& 99.8\%
\\\cline{2-4} & Depth &94.55\%&88.3 \%
\\\cline{2-4}&Fusion &98.40\%& 99.4 \% \\ 
\hline
\hline

Yaw $\pm 60$\degree & RGB &93.75 \%& 97.4\%
\\\cline{2-4} & Depth  & 86.05\%&87.0\%
\\\cline{2-4}&Fusion & 93.43\%& 98.2\% \\ 
\hline
\hline

Yaw $\pm 90$\degree & RGB & 70.2\%& 83.7\%
\\\cline{2-4} & Depth & 63.45\%& 74\%
\\\cline{2-4}&Fusion & 71.15\%&84.6 \%\\ 
\hline
\hline
Average & RGB& 94.6\%& 96.70\%
\\\cline{2-4}  & Depth& 88.67\%& 86.65\%
\\\cline{2-4} & Fusion & 94.23\% & 96.98\%\\ 
\hline
\end{tabular} 
}
\end{table}
\\
The presented results demonstrates that our method can works equally to the aforementioned work which is based on expensive pre-processing stage to frontalize and to fill symmetrically the self-occluded part in the face due to head rotation. An overall of 94.6\%, 88.67\% and 94.23\% are achieved with our system respectively for RGB, depth and multimodal data. It's clear that our RGB performance need to be improved but our depth performance seems better than this reported in \cite{li2013using} which demonstrate the importance and the effectiveness of data representation with learning discriminant features to overcome challenging conditions.
In other hand we present in table \ref{tab:perf_curtinface} our obtained results in comparison with the most performing state of the art techniques namely\cite{ciaccio2013face,hsu2014rgb} and some recent works \cite{grati2016scalable,kaashki2018rgb}.

It is shown that our system outperforms \cite{ciaccio2013face} (Cov+LBP) with a margin of 4 \% in Yaw $\pm 30 $ while a gain of $\approx 9\%$ in Yaw $\pm 60 $. The drop in the performance for the set Yaw $\pm 90$ angles can be explained by the fact that CurtinFaces database contains only just two samples for left and right $\pm 90$. That is, if we take one sample for testing, no corresponding pose exists anymore in the gallery. In contrary and as reported previously in the related work section, \cite{ciaccio2013face,hsu2014rgb,li2013using} pre-processings allow to tackle this issue as they either augment the gallery by generating new poses, correct the pose or symmetrically filling the self occluded part in the face. Beyond these well engineered pre-processings, in this work we aim to focus on estimating optimal RGB-D data representation for face recognition applications. 

In other hand, our study is compared also to \cite{kaashki2018rgb} (labeled as HLF+SVM in the Table \ref{tab:perf_curtinface}) as it uses patch representation around landmarks points. We can observe that our performance are better with a gain of 8\% in Yaw $\pm 30$ angle and an improvement of more than 30\% in Yaw $\pm 60$ angle. These results highlights the added-value of learned features to derive more discriminant representations for local features compared to the standard hand-crafted features (\ie HOG, LBP, and 3DLBP) and prove clearly the use of salient points without interpreting face structure.

On Eurecom database, the first session set is selected for training and the second one for test. The learned feature descriptor yield to 90,82\% for texture images and 85,57\% for depth data, and 92,70\% after fusion, which is better than the recognition rate obtained in RISE \cite{goswami2014rgb} (\ie 89.0\% after fusion). 

The second set of experiments are dedicated to compare our CNN learned features to hand-crafted features taking our system as baseline. This is to highlight the added-value of CNN learned features. Three versions of our system are tested, the only changed part is the feature extraction: HOG+SRC, LBP+SRC, and CNN+SRC. As shown in Figure \ref{fig:handcraftedvsLearned}, CNN+SRC outperforms the other systems for all the test subsets.

Based on all the obtained results and comparisons, we can conclude that CNN learned-features can achieve a competitive identification performance for person recognition from low-cost sensor and under challenging pose and expression variations and without any prior face analysis (\eg face pose estimation, facial landmarks detection, \etc).

\section{\uppercase{Conclusion}}
\label{sec:conclusion}
In this paper, we proposed a data-driven representation for RGB-D face recognition. A given face is represented by a set of patches around detected salients key-points on which a CNNs transformation are applied to extract the learned local descriptor for each modality separately before performing SRC classification. The experimental results obtained on benchmark RGB-D databases highlight the added-value of deep learning local features compared to standard hand-crafted feature extractors. For future works, we plan to extend our system with learning a multi-modal representation to combine texture and depth data. With an appropriate CNN, the fusion strategy of RGB-D data will take in consideration the complementarity between depth and image data to enhance the recognition performance. 

\begin{table*}[h]
\caption{CurtinFaces Database performances on differently approach.}\label{tab:example1} \centering
\label{tab:perf_curtinface}
\begin{tabular}{lllllll}

Pose  &   Our &  Cov+LBP & LBP+SRC & DCS+SRC   & BSIF+SRC    & HLF+SVM\\
\hline
Frontal   & 100\% & N/A     & 100\%    & 100\% & 100\% & 100\%\\
\hline
Yaw $\pm 30$  & 98.40\%  & 94.2\%  & 99.4 \%  & 99.8\%  & 99.04\% & 90.3\%\\
\hline
Yaw $\pm 60$ &  93.43\% &   84.6\%  & 98.2 \%  & 97.4 & 95.51\% & 58.6\% \\
\hline
Yaw $\pm 90$ & 71.15\% &  75.0\%   & 93.5\% & 83.7\% & 60.58\% & N/A \\
\hline
\end{tabular}
\end{table*}


\vfill
\end{document}